\documentclass{article}
\usepackage{graphicx} 
\usepackage{twocolceurws}

\usepackage[T1]{fontenc}

\usepackage[
  backend=biber,
  style=numeric-comp,
  sorting=none
]{biblatex}

\usepackage{svg}

\title{Integrating Generative AI into Art Therapy: A Technical Showcase}
\author{
Yannis Valentin Schmutz \\ yannis.schmutz@bfh.ch
\and
Tetiana Kravchenko \\ tetiana.kravchenko@powercoders.org\\
\and
Souhir Ben Souissi \\ souhir.bensouissi@bfh.ch
\and
Mascha Kurpicz-Briki \\ mascha.kurpicz@bfh.ch 
}
\date{July 2024}

\institution{Generative AI Lab \\Bern University of Applied Sciences}

\begin{document}
\maketitle

\begin{abstract}
This paper explores the integration of generative AI into the field of art therapy.
Leveraging proven text-to-image models, we introduce a novel technical design to complement art therapy. 
The resulting AI-based tools shall enable patients to refine and customize their creative work, opening up new avenues of expression and accessibility. 
Using three illustrative examples, we demonstrate potential outputs of our solution and evaluate them qualitatively. 
Furthermore, we discuss the current limitations and ethical considerations associated with this integration and provide an outlook into future research efforts. 
Our implementations are publicly available at https://github.com/BFH-AMI/sds24.
\end{abstract}

\section{Introduction}

Mental health is a crucial factor for general well-being \cite{prince2007} and the productivity of individuals \cite{oliveria2023}.
Mental health problems such as stress, depression and anxiety are widespread \cite{michel2024, mehmeti2023, richter2019} and pose a major challenge to both personal and public health \cite{arbues2020, moitra2023}.
For several years psychotherapy has proven to be an effective treatment for these disorders \cite{lambert1994}.
A special form of psychotherapy is art therapy, in which artistic media are used as the primary means of expression \cite{BritAssoc2015}.
In this type of treatment, creative activities such as drawing, sculpting and making music are used to promote psychological healing \cite{bitonte2014}. 
These non-verbal activities help patients to express their feelings and explore their thoughts \cite{haeyen2018} while, among others, relieving stress and processing traumatic events \cite{haeyen2022, hu2021}.

With the advancement of digital technologies, the effectiveness of psychotherapy can be even further improved \cite{berger2018}.
In particular, innovations in the intersection between artificial intelligence and mental health have the potential to improve global accessibility \cite{fiske2019}, engagement \cite{spiegel2024} and effectiveness \cite{creswell2018}.
While psychotherapy primarily allows applications from the field of natural language generation \cite{das2022, khalaf2023, montag2024}, art therapy offers further possibilities with other modalities.  
However, the application of AI technologies in digital art therapy is still under-explored \cite{liu2024, du2024}.
A quick search on PubMed with the query \textit{"artificial intelligence"[Title/Abstract] AND "art therapy"[Title/Abstract]} leads to only two results. 
Kwon et al. \cite{kwon2024} discuss the future challenges of combining five-element music therapy and AI.
Kim et al. \cite{kim2023} reveal the feasibility of creating an evaluation dataset for digital integrative art therapy content for the treatment of depression.
Only one other relevant work \cite{10.1145/3689649} \cite{10.1145/3581641.3584087} was identified in the ACM Digital Library: The authors use an explainable AI-based expert system to support the process of assessing drawings from art therapy for clinical professionals.

These limited results illustrate that research at the interface between generative AI and art therapy is still at an early stage. Therefore, there is a considerable need for further studies investigating technical approaches to combining AI and art therapy and the associated challenges.
In particular, this concerns ethical aspects, especially in relation to data protection, integration and responsible use.
Addressing these issues requires continuous attention and proactive measures.

In this article, we showcase a potential technical design that combines art therapy and generative AI.
Our approach is by no means intended to replace art therapy in general or the therapist in any way. Rather, we want to show how generative AI can support accessibility and expressiveness. We target a technology in the sense of an \emph{augmented intelligence} supporting human domain experts instead of an \emph{artificial intelligence} aiming to replace them. The technology serves as a tool to support the human processes, in this case in the context of art therapy. By using not only physical but also virtual art methodologies, the existing art therapy can be extended by additional components. 
Furthermore, we discuss the current technical limitations as well as ethical concerns and show directions how these can be mitigated by future research.

\section{Methodology}
We start this section by defining a simplified process of an art therapy session. 
We then show how it could be supplemented by integrating generative AI models to the process.

\subsection{Therapeutic Process}
\label{sec:therapeutic_process}
In practice, the actual application of art therapy differs depending on the therapist, patient and their condition. Considering recent sources \cite{elswick2019, miloszewska2020}, we summarize the process of one therapy session in four simplified exemplary phases and limit it to the therapeutic form of painting. 
In the first step, the \textit{(i) initial conversation}, the patient names the current problem they are dealing with in their life. Together with the therapist, the associated emotions, themes and possible types of visualization are explored. Based on this, \textit{(ii) the artistic work} is created. During this process, the patient mindfully observes his feelings and expresses them artistically. The therapist accompanies this process by asking specific questions and helps the patient to stay on topic. 
Once the patient has reached a suitable point, the painting is paused and the \textit{(iii) adaptation phase} is initiated. In an interim discussion, the therapist examines with the patient what is going on in the picture, which aspects are included and which are not. 
Together they try to find out what would be beneficial to the topic or which parts of the drawing need adaption.
The patient then implements this through creative adjustments such as painting over individual parts of the picture.
This is followed by \textit{(iv) the retrospective conversation}, during which they discuss what the adjustments made mean for the patient's daily life and what creative solutions to the current problem might look like. 

\begin{figure}[ht!]
  \includegraphics[width=0.48\textwidth]{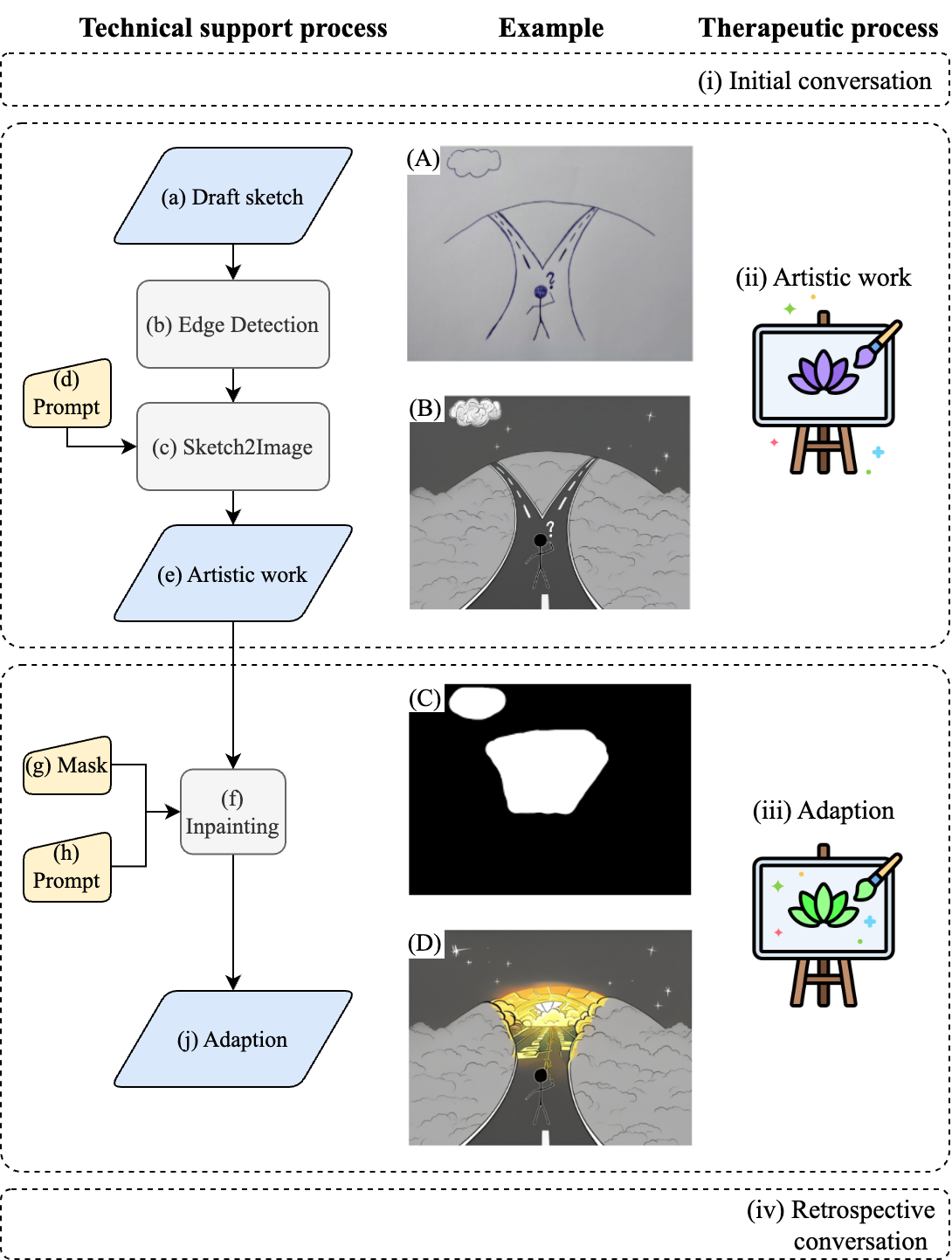}
  \centering
  \caption{Inclusion of generative AI in the art therapy process.}
    \label{fig:process}
\end{figure}

\subsection{Technical Support Process}

Our technical process supports the phases (ii) and (iii) of the therapeutic process presented in Section \ref{sec:therapeutic_process}, the creation and adaptation of the artistic work.
The use of generative AI makes it possible to further refine a drawing by means of text descriptions and even to recreate entire sections.
This potentially allows patients to express themselves further and in additional ways.
Figure \ref{fig:process} shows the phases of the therapeutic process top-to-bottom as well as the associated technical components.
The middle column shows examples of the individual steps.
We illustrate our process with a fictitious patient who is metaphorically at a crossroads in his life. 
This causes him to feel insecure and anxious.
\\
After the initial conversation, the client draws a first draft (A) that visualizes their current problem and emotional state.
For the practical application of our process, the drawing must be available in a digital form. This can be achieved by the patient working directly on a tablet or alternatively by scanning or photographing the paper drawing.
Next, the patient examines together with the therapist which aspects may not yet be included in the picture. 
These are noted in the form of keywords and input into a text-to-image model together with the digitized draft.
This then results in the refined artistic work (B).
In the example shown, the sketch was supplemented by a text description that describes a gloomy scene and the client's current insecurity.
This step can be repeated as often as required until the patient is satisfied with the result.
We then move on to phase (iii) by examining which adjustments to the picture could be beneficial. 
In our example, the patient desires a clear path forward, out of the gloomy mood and into a bright environment that gives him clarity. 
Accordingly, he marks the section of the picture he wants to change and describes his intention. 
The image (B) and marking (C) are passed together with the description to the inpainting model, which outputs the adapted image (D).
In the retrospective conversation (iv), the patient then examines with his therapist how he could achieve the desired state, following the same procedures as in traditional art therapy. 

Technically, we implemented this process using three deep learning models, as shown in Figure \ref{fig:process} on the left. 
Our solution only uses models that can be run locally on standard graphics cards and do not require API calls to third parties. This allows us to ensure the privacy of sensitive data from the ground up.
\\
We start the technical support process by detecting and extracting the edges (b) of the drawn draft (a) using PiDiNet \cite{su2021}.
These are then input into a sketch-to-image model (c) \cite{mou2024sketch} together with the text prompt (d). 
Specifically, this is a text-to-image adapter \cite{mou2024}, which enables further conditioning of existing T2I models.
In this case, we use the adapter to align the output of a stable diffusion model \cite{podell2023} to the detected edges.
In doing so, the client's drawing remains decisive for the generated output (e).
To perform the adaptation of the image, we use inpainting (f). Inpainting describes the procedure of filling missing, i.e. masked, parts of an image with credible content. Our chosen model \cite{kandinsky2023} further allows to condition the replacement content based on a text description (h). 
The area to be replaced is defined by a binary mask (g), which is placed over the image by the user.
In our example, we let the model replace the cloud, the two roads and the question mark. 
All unmarked sections remain unchanged.
The resulting adapted image (j) may then serve as basis for the retrospective conversation (iv).

\section{Results}

\begin{figure}[ht!]
  \includegraphics[width=0.48\textwidth]{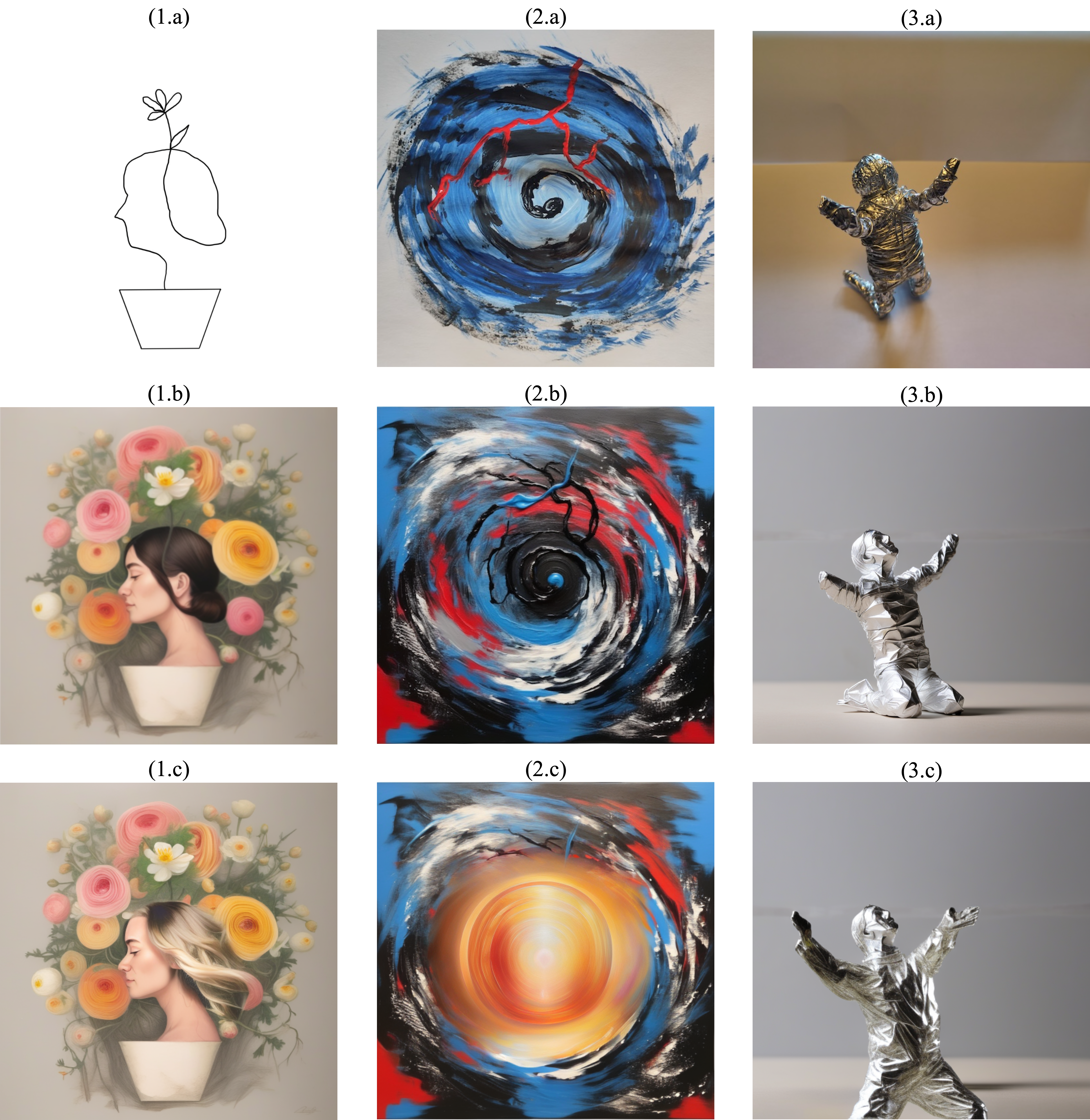}
  \centering
  \caption{Example outputs of the employed generative AI models}
  \label{fig:showcase}
\end{figure}

We qualitatively evaluated the applicability of the chosen models using three different image inputs characterized by increasing visual complexity.
The image inputs include: (1) a black and white sketch created on a tablet, (2) a picture painted by us with acrylic paint, and (3) a photo of a custom-made 3D figure made of aluminum foil. 
Figure \ref{fig:showcase} showcases the drafts, artistic works and the adaptations in rows (a), (b), and (c), respectively. 
The corresponding prompts are listed in Table \ref{tab:prompts}.
In accordance with our methodology, the artwork (1.b) was created on the basis of the draft (1.a) and the prompt (P1.b). The adapted image (1.c) resulted from the inputs (1.b) and (P1.C). The same principle applies to the other examples.
The generation took between 5 and 10 seconds per image on an NVIDIA RTX A6000 graphics card.

\begin{table}[]
    \begin{tabular}{|c|p{0.375\textwidth}|}
    \hline
    \textbf{ID} & \textbf{Prompt} \\ \hline
    (P1.b) & A head of woman in pot, realistic facial features, ranunculus on her head, representing personal growth and renewal, messy background, detailed, high quality. \\ \hline
    (P1.c) & A woman with a long, flowing blonde hairstyle, adding waves and volume. \\ \hline
    (P2.b) & An abstract painting showing a spiral. It symbolises insecurity and fears. Blue and black colors. There's a red thunderstorm shaped line from top to the center of the spiral. \\ \hline
    (P2.c) & A painting of light spreading spherically from the center, warm colors, representing calmness, clarity and hope, matching the artistic style. \\ \hline
    (P3.b) & A small figure made of aluminum foil and wire depicting a man on his knees with his arms up in the air. The man seems helplessness. \\ \hline
    (P3.c) & A small figure made of aluminum foil depicting a man in an upright position with his arms up in the air. \\ \hline
    \end{tabular}
    \caption{Prompts used to create example images.}
    \label{tab:prompts}
\end{table}

The first draft (1.a) forms the simplest example, as it solely consists of uniform black lines on white background. 
It shows a silhouette of a woman's head growing out of a pot, drawn with a single, continuous line.
Despite the simplicity of the image, the model is able to capture its essence and generate an artistic work that fits the prompt (P1.b) very well.
The same is true for its adaptation (1.c).
\\
The second draft (2.a) presents a higher level of complexity compared to the initial example, primarily due to its abstract and less defined shapes and edges. 
This increased vagueness may challenge the model's ability to accurately interpret and generate the intended visual elements.
We see this partly reflected in the output (2.b). On one hand, the image is entirely filled with colors and no longer forms a circle on a white background as in the original. On the other hand, the red, lightning-shaped lines have become diffused throughout the entire artistic work. 
This outcome is likely also influenced by its somewhat imprecise prompt (P2.b).
The situation improves with the adaptation (2.c). Here, the adjustments in the image are closely aligned with the prompt (P2.c), while the artistic style remains intact and undistorted.
\\
So far we have only used drawings as input. The last example (3.a) is intended to demonstrate the possibility of including photos of any visual artwork, such as a sculpture, in the process. For this, we formed a figure of a person kneeling and raising their hands in the air out of aluminum foil. We believe that a photo of a 3D object represents the greatest challenge for the generative model used. After all, essential aspects of the work are lost through photography and the associated reduction to two dimensions.
However, as can be seen in the output (3.b), the essence of the sculpture has essentially been adopted. Only the missing hands stand out negatively, although these are not depicted in the original figure either.
Through the adaptation, the kneeling man could be brought into an upright position by masking and replacing the whole body, apart from the head. The output (3.c) aligns well with the prompt (P3.c) and now also features simple hands.

\section{Discussion}

This preliminary study has shown the potential of generative AI to provide new dimensions in the context of art therapy. By combining existing and approved methods with novel technology, new approaches are possible for the clinical professionals and their patients or clients. We have showcased a potential architecture with some examples, which can serve as a foundation for future research. However, there are some limitations in our approach which are detailed in the following subsection. We will then at the end of the section provide an outlook for this field of research and potential application use cases.

\subsection{Limitations}

The process presented is merely a proof-of-concept for a possible approach to combine generative AI and art therapy. There are therefore still a number of limitations and challenges that need to be addressed before the actual effectiveness of such a combination can be investigated. We will discuss these in more detail in the following. 

\begin{itemize}
\renewcommand{\labelitemi}{}
\item \textit{Edge Detection Accuracy:} The quality of the edge detection strongly impacts the output of the sketch-to-image adapter. If the edges are not accurately detected, the generated image might not align well with the patient's draft, leading to a general loss of the patient's intention.
Furthermore, any information about the colors used is lost in this step. If these cannot be implicitly concluded from the shapes of the edges, they must be specified in the text description.

\item \textit{Model Generalization:} The image-to-image model might struggle with generalizing across different artistic styles and varying quality of the painting. This could lead to less suitable results, especially if the sketches entered are rough or abstract.

\item \textit{Inpainting Quality:} Inpainting models may have difficulty seamlessly integrating new content into the existing image, especially if the area to be replaced is large or complex. This can lead to visual artifacts or inconsistencies in the final image.

\item \textit{Therapeutic Efficacy:} Given the novelty of these methods, a formal assessment of their therapeutic efficacy in form of clinical trials is required, as it has been done for traditional art therapy (e.g., Abbing et al. \cite{abbing2019}).

\item \textit{User Acceptance:} Additionally, it needs to be assessed if, and in what form, this novel variant of art therapy is acceptable to the users, both patients/clients and health professionals. However, at the same time, since the aim is to extend previous methods with additional options, this gives also potential to include additional patients/clients not previously using this type of therapy.

\item \textit{Bias and Content Appropriateness:} The stereotypes of the society are known to be included in AI models. This is in particular also the case for image generation (e.g., \cite{chauhan2024}). This can lead to a reproduction of stereotypes as well as inappropriate content. This needs to be considered when implemented systems for a real-world use, especially that the therapeutic environment should be considered a safe space. 

\end{itemize}

\subsection{Outlook \& Potential Future Applications}
We have introduced an initial proof of concept for an approach that merges generative AI with art therapy.
Although there are some limitations, as previously discussed, this approach offers promising potential across various dimensions, suggesting new pathways for therapeutic innovation and patient engagement.
\begin{itemize}
\renewcommand{\labelitemi}{}
\item \textit{Virtual Art Therapy:} The adaptable nature of art therapy allows for its implementation on virtual platforms, facilitating remote sessions \cite{feen2023creating}. Through digital painting tools and virtual reality (VR) environments, patients can create art and connect with therapists from afar, enhancing the accessibility of therapy \cite{feen2023creating}. In addition, VR art creation offers unique features like three-dimensional painting and dynamic scaling, providing an immersive and expressive creative experience. A critical element of VR in psychotherapy is its capacity to evoke a strong sense of "presence" within the computer-generated environment \cite{riva2016transforming}. This is achieved by accurately simulating sensory (such as visual and auditory) and motor cues (including immersive environments and motion tracking) similar to those in the real world. Such immersive experiences can significantly enhance therapeutic outcomes by fostering a sense of reality, encouraging personal growth and introspection by allowing individuals to "experience" their transformation \cite{hacmun2018principles}.
\item \textit{Combination with Other Therapies:} 
In clinical settings, our innovative approach that integrates generative AI with art therapy is well-suited for combination with other therapeutic modalities, such as cognitive-behavioral therapy (CBT) or mindfulness practices. For example, following an initial adaptation phase, therapists might employ CBT methods to strengthen positive shifts in a patient's thought processes \cite{rosal2018cognitive}. This integration is particularly relevant in cognitive-behavioral art therapy (CBAT), which is now an established part of treatment protocols for conditions like post-traumatic stress disorder (PTSD) \cite{rosal2015cognitive, rosal2016cognitive}. Furthermore, in collaboration with neuroscientists, future research could investigate how various phases of art therapy influence brain activity. This would allow therapists to adjust their methods based on neurofeedback, potentially increasing the efficacy of treatments tailored to individual brain responses \cite{ordikhani2016neurofeedback, morrissey2024could}. 
\item \textit{Extended Therapeutic Applications:}
Our approach can be applied to long-term management of chronic conditions like depression or anxiety. Each phase can be revisited periodically to reflect changes in the patient's life circumstances \cite{hamre2007anthroposophic, hu2021art}.
Furthermore, this approach can be adapted for patients with dementia or other cognitive impairments. In these cases, the focus shifts towards maintaining cognitive functions and enhancing quality of life through creative expression \cite{masika2020visual}. Integrating Generative AI into Art Therapy in this context will not only help in preserving cognitive abilities \cite{lee2019art} but may also contribute to slowing cognitive decline \cite{mahendran2018art}.
Additionally, this combination of Generative AI and Art Therapy holds potential for early detection of dementia. Early detection is crucial for preserving the quality of life for patients and can significantly reduce the costs associated with long-term care \cite{chen2020automatic,jiang2021mobile}.
\end{itemize}

\printbibliography
\end{document}